%% file: rothkopf_95.tex
\documentclass[runningheads,a4paper]{llncs}

\include{preamble}
\usepackage{psfrag}

\usepackage{url}

\newcommand{\keywords}[1]{\par\addvspace\baselineskip
\noindent\keywordname\enspace\ignorespaces#1}

\begin{document}

\title{Preference elicitation and\\ inverse reinforcement learning}
\author{Constantin A. Rothkopf\inst{1} \and Christos Dimitrakakis\inst{2}}
\institute{Frankfurt Institute for Advanced Studies, Frankfurt, Germany\\
  \email{rothkfopf@fias.uni-frankfurt.de}
  \and
  EPFL, Lausanne, Switzerland\\
  \email{christos.dimitrakakis@epfl.ch} 
}

\toctitle{Preference elicitation and inverse reinforcement learning}
\tocauthor{C. Rothkopf and C. Dimitrakakis}

\maketitle

\begin{abstract}
  We state the problem of inverse reinforcement learning in terms of preference elicitation, resulting in a principled (Bayesian) statistical formulation. This generalises previous work on Bayesian inverse reinforcement learning and allows us to obtain a posterior distribution on the agent's preferences, policy and optionally, the obtained reward sequence, from observations. We examine the relation of the resulting approach to other statistical methods for inverse reinforcement learning via analysis and experimental results. We show that preferences can be determined accurately, even if the observed agent's policy is sub-optimal with respect to its own preferences. In that case, significantly improved policies with respect to the agent's preferences are obtained, compared to both other methods and to the performance of the demonstrated policy.
\end{abstract}

\section{Introduction}

Preference elicitation is a well-known problem in statistical decision
theory~\citep{friedman1952euh}. The goal is to determine, whether a
given decision maker prefers some events to other events, and if so,
by how much. The first main assumption is that there exists a partial
ordering among events, indicating relative preferences. Then the
corresponding problem is to determine which events are preferred to
which others. The second main assumption is the expected utility
hypothesis. This posits that if we can assign a numerical {\em
  utility} to each event, such that events with larger utilities are
preferred, then the decision maker's preferred choice from a set of
possible {\em gambles} will be the gamble with the highest {\em
  expected} utility. The corresponding problem is to determine the
numerical utilities for a given decision maker.

Preference elicitation is also of relevance to cognitive science and
behavioural psychology, e.g. for determining rewards implicit in
behaviour~\cite{Rothkopf2008PhD} where a proper elicitation procedure
may allow one to reach more robust experimental conclusions. There are
also direct practical applications, such as user modelling for
determining customer preferences~\cite{boutilier2002pomdp}.  Finally,
by analysing the apparent preferences of an expert while performing a
particular task, we may be able to discover behaviours that match or
even surpass the performance of the
expert~\cite{abbeel2004apprenticeship} in the very same task.

This paper uses the formal setting of preference elicitation to
determine the preferences of an agent acting within a discrete-time
stochastic environment. We assume that the agent obtains a sequence of
(hidden to us) rewards from the environment and that its preferences
have a functional form related to the rewards. We also suppose that
the agent is acting nearly optimally (in a manner to be made more
rigorous later) with respect to its preferences.  Armed with this
information, and observations from the agent's interaction with the
environment, we can determine the agent's preferences and policy in a
Bayesian framework. This allows us to generalise previous Bayesian
approaches to inverse reinforcement learning.

In order to do so, we define a structured prior on reward functions
and policies. We then derive two different Markov chain procedures for
preference elicitation. The result of the inference is used to obtain
policies that are significantly improved with respect to the {\em true
  preferences} of the observed agent. We show that this can be
achieved even with fairly generic sampling approaches.

Numerous other inverse reinforcement learning approaches
exist~\cite{abbeel2004apprenticeship,ramachandran51bayesian,syed-schapire:game-theory:nips,apprenticeship:classification}.
Our main contribution is to provide a clear Bayesian formulation of
inverse reinforcement learning as preference elicitation, with a
structured prior on the agent's utilities and policies. This
generalises the approach of~\citet{ramachandran51bayesian} and paves
the way to principled procedures for determining distributions on
reward functions, policies and reward sequences. Performance-wise, we
show that the policies obtained through our methodology easily surpass
the agent's actual policy with respect to its own
utility. Furthermore, we obtain policies that are significantly better
than those obtained with other inverse reinforcement learning methods
that we compare against.

Finally, the relation to {\em experimental design} for preference
elicitation (see \cite{boutilier2002pomdp} for example) must be
pointed out. Although this is a very interesting planning problem, in
this paper we do not deal with {\em active} preference elicitation. We
focus on the sub-problem of estimating preferences given a particular
observed behaviour in a given environment and use decision theoretic
formalisms to derive efficient procedures for inverse reinforcement
learning.

This paper is organised as follows.  The next section formalises the
preference elicitation setting and relates it to inverse reinforcement
learning. Section~\ref{sec:statistical-model} presents the abstract
statistical model used for estimating the agent's
preferences. Section~\ref{sec:model-estimation} describes a model and
inference procedure for joint estimation of the agent's preferences
and its policy.  Section~\ref{sec:related-work} discusses related work
in more detail.  Section~\ref{sec:experiments} presents comparative
experiments, which quantitatively examine the quality of the solutions
in terms of both preference elicitation and the estimation of improved
policies, concluding with a view to further extensions.

\section{Formalisation of the problem}
\label{sec:formalisation}
We separate the agent's preferences (which are unknown to us) from the
environment's dynamics (which we consider known).  More specifically,
 the environment is a controlled Markov process $\cmp = (\States, \Actions,
\Trans)$, with state space $\States$, action space $\Actions$, and
transition kernel $\Trans = \cset{\tran{\cdot}{s,a}}{s \in \States, a
  \in \Actions}$, indexed in $\States \times \Actions$ such that
$\tran{\cdot}{s,a}$ is a probability measure\footnote{We assume the
  measurability of all sets with respect to some appropriate
  $\sigma$-algebra.} on $\States$. The dynamics of the environment are
Markovian: If at time $t$ the environment is in state $s_t \in S$ and
the agent performs action $a_t \in A$, then the next state $s_{t+1}$
is drawn with a probability independent of previous states and
actions:
\begin{equation}
  \label{eq:next-state}
  \Pr_\cmp(s_{t+1} \in S \mid s^t, a^t) = \tran{S}{s_t,a_t},
  \qquad
  S \subset \States,
\end{equation}
where we use the convention $s^t \equiv s_1, \ldots, s_t$ and $a^t
\equiv a_1, \ldots, a_t$ to represent sequences of variables.

In our setting, we have observed the agent acting in the environment
and obtain a sequence of actions and a sequence of states:
\begin{align*}
  D &\defn (a^T, s^T),
  &
  a^T &\equiv a_1, \ldots, a_T,
  &
  s^T &\equiv s_1, \ldots, s_T.
\end{align*}
The agent has an {\em unknown utility function}, $U_t$, according to
which it selects actions, which we wish to discover. Here, we assume
that $U_t$ has a structure corresponding to that of reinforcement
learning infinite-horizon discounted reward problems and that the
agent tries to maximise the expected utility.
\begin{assumption}
  The agent's utility at time $t$ is the total $\disc$-discounted
  return from time $t$:
  \begin{equation}
    U_t \defn \sum_{k=t}^\infty \disc^k r_k,
  \end{equation}
  where $\disc \in [0,1]$ is a discount factor, and the reward $r_t$
  is given by the (stochastic) reward function $\rew$ so that $r_t
  \mid s_t = s, a_t = a \sim \rew(\cdot \mid s, a)$, $(s, a) \in
  \States \times \Actions$.
\end{assumption}
This choice establishes correspondence with the standard reinforcement
learning setting.\footnote{In our framework, this is only one of the
  many possible assumptions regarding the form of the utility
  function.  As an alternative example, consider an agent who collects
  gold coins in a maze with traps, and with a utility equal to the
  logarithm of the number of coins if it exists the maze, and zero
  otherwise.} The controlled Markov process and the utility define a
Markov decision process~\citep{Puterman:MDP:1994} (MDP), denoted by
$\mdp = (\States, \Actions, \Trans, \rew, \disc)$.  The agent uses
some policy $\pol$ to select actions with distribution $\pol(a_t \mid
s_t)$, which together with the Markov decision process $\mdp$ defines
a Markov chain on the sequence of states, such that:
\begin{align}
  \Pr_{\mdp,\pol}(s_{t+1} \in S \mid s^t)
  &=
  \int_\Actions \tran{S}{a,s_t} \dd{\pi}(a \mid s_t),  
\end{align}
where we use a subscript to denote that the probability is taken with
respect to the process defined jointly by $\mdp, \pol$. We shall use
this notational convention throughout this paper.  Similarly, the {\em
  expected utility} of a policy $\pol$ is denoted by $\E_{\mdp,\pol}
U_t$.  We also introduce the family of $Q$-value functions
$\cset{\Qf^\pol}{\mdp \in \MDPs, \pol \in \Pols}$, where $\MDPs$ is a
set of MDPs, with $\Qfp : \States \times \Actions \to \Reals$ such
that:
\begin{equation}
  \label{eq:q-function}
  \Qf^\pol(s, a) \defn \E_{\mdp, \pol} \left(U_t \mid s_t = s, a_t = a\right).
\end{equation}
Finally, we use $\Qfs$ to denote the optimal $Q$-value function for an
MDP $\mdp$, such that:
\begin{equation}
  \label{eq:optimal-q-function}
  \Qfs(s,a) = \sup_{\pol \in \Pols} \Qfp(s,a),
  \qquad
  \forall s \in \States, a \in \Actions.
\end{equation}
With a slight abuse of notation, we shall use $Q_\rew$ when we only
need to distinguish between different reward functions $\rew$, as long as
the remaining components of $\mdp$ remain fixed.

Loosely speaking, our problem is to estimate the reward function
$\rew$ and discount factor $\disc$ that the agent uses, given the
observations $s^T, a^T$ and some prior beliefs. As shall be seen in
the sequel, this task is easier with additional assumptions on the
structural form of the policy $\pol$.  We derive two sampling
algorithms. The first estimates a joint posterior distribution on the
policy and reward function, while the second also estimates a
distribution on the sequence of rewards that the agent obtains.  We
then show how to use those estimates in order to obtain a policy that
can perform significantly better than that of the agent's original
policy with respect to the agent's true preferences.

\section{The statistical model}
\label{sec:statistical-model}
In the simplest version of the problem, we assume that $\disc, \cmp$
are known and we only estimate the reward function, given some prior
over reward functions and policies. This assumption can be easily
relaxed, via an additional prior on the discount factor $\disc$ and
CMP $\cmp$.  Let $\Rews$ be a space of reward functions $\rew$ and
$\Pols$ to be a space of policies $\pol$. We define a (prior)
probability measure $\rbel(\cdot \mid \cmp)$ on $\Rews$ such that for
any $B \subset \Rews$, $\rbel(B \mid \cmp)$ corresponds to our prior
belief that the reward function is in $B$.  Finally, for any reward
function $\rew \in \Rews$, we define a conditional probability measure
$\pbel(\cdot \mid \rew, \cmp)$ on the space of policies $\Pols$.  Let
$\agrew, \agpol$ denote the agent's true reward function and policy
respectively.  The joint prior on reward functions and policies is denoted by:
\begin{align}
    \jbel(P, R \mid \cmp)
    &\defn
    \int_R \pbel(P \mid \rew, \cmp) \dd{\rbel}(\rew \mid \cmp),
    &
    P &\subset \Pols,
    R \subset \Rews,
  \label{eq:joint-prior}
\end{align}
such that $\jbel(\cdot \mid \cmp)$ is a probability measure on $\Rews
\times \Pols$.
\begin{figure}[t]
  \centering
  \subfigure[Basic model]{
    \hspace{2em}
    \begin{tikzpicture}[x=0.5\linewidth]
      \node[observed] (RewardPrior) {$\rbel$};
      \node[observed] (PolicyPrior) [below of=RewardPrior] {$\pbel$};
      \node[hidden] (Reward) [right of=RewardPrior] {$\rew$}; 
      \node[hidden] (Policy) [right of=PolicyPrior]{$\pol$};
      \node[observed] (Data) [right of=Reward] {$D$};
      \draw[->] (RewardPrior) -- (Reward);
      \draw[->] (Reward) -- (Policy);
      \draw[->] (PolicyPrior) -- (Policy);
      \draw[->] (Policy) -- (Data);
    \end{tikzpicture}
    \hspace{2em}
    \label{fig:basic-model}
  }
  \hspace{1em}
  \subfigure[Reward-augmented model]{
    \hspace{2em}
    \begin{tikzpicture}[x=0.5\linewidth]
      \node[observed] (RewardPrior) {$\rbel$};
      \node[observed] (PolicyPrior) [below of=RewardPrior] {$\pbel$};
      \node[hidden] (Reward) [right of=RewardPrior] {$\rew$}; 
      \node[hidden] (Policy) [right of=PolicyPrior]{$\pol$};
      \node[hidden] (Rewards) [right of=Policy]{$r^{T}$};
      \node[observed] (Data) [right of=Reward] {$D$};
      \draw[->] (RewardPrior) -- (Reward);
      \draw[->] (Reward) -- (Rewards);
      \draw[->] (PolicyPrior) -- (Policy);
      \draw[->] (Policy) -- (Data);
      \draw[->] (Data) -- (Rewards);
      \draw[->] (Reward) -- (Policy);
    \end{tikzpicture}
    \hspace{2em}
    \label{fig:augmented-model}
  }
  \caption{Graphical model, with reward priors $\rbel$ and policy
    priors $\pbel$, while $\rew$ and $\pol$ are the reward and policy,
    where we observe the demonstration $D$. Dark colours denote
    observed variables and light denote latent variables.  The
    implicit dependencies on $\cmp$ are omitted for clarity.}
\label{fig:model}
\end{figure}
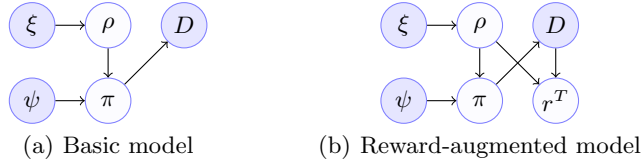
We define two models, depicted in Figure~\ref{fig:model}.  The {\em
  basic model}, shown in Figure~\ref{fig:basic-model}, is defined as
follows:
  \begin{align*}
    \rew & \sim \rbel(\cdot \mid \cmp),&
    \pol \mid \agrew = \rew & \sim \pbel(\cdot \mid \rew, \cmp),
  \end{align*}

We also introduce a {\em reward-augmented model}, where we explicitly
model the rewards obtained by the agent, as shown in
Figure~\ref{fig:augmented-model}:
\begin{align*}
  \rew & \sim \rbel(\cdot \mid \cmp),
  &
  \pol \mid \agrew = \rew & \sim \pbel(\cdot \mid \rew, \cmp),
  &
  r_t \mid \agrew = \rew, s_t = s, a_t = a & \sim \rew(\cdot \mid s, a).
\end{align*}

For the moment we shall leave the exact functional form of the prior
on the reward functions and the conditional prior on the policy
unspecified.  Nevertheless, the structure allows us to state the
following:
\begin{lemma}
  For a prior of the form specified in \eqref{eq:joint-prior}, and
  given a controlled Markov process $\cmp$ and observed state and
  action sequences $s^T, a^T$, where the actions are drawn from a
  reactive policy $\pol$, the posterior measure on reward functions
  is:
  \begin{equation}
    \label{eq:reward-function-posterior}
      \rbel(B | s^T, a^T, \cmp)
      =
      \frac{\int_B
      \int_\Pols
      \pol(a^T | s^T)
      \dd{\pbel}(\pol | \rew, \cmp)
      \dd{\rbel}(\rew | \cmp)
    }
    {
      \int_\Rews
      \int_\Pols
      \pol(a^T | s^T)
      \dd{\pbel}(\pol | \rew, \cmp)
      \dd{\rbel}(\rew | \cmp)
    },
  \end{equation}
  where $\pol(a^T \mid s^T) = \prod_{t=1}^T \pol(a_t | s_t)$.
\label{lem:prior-lemma}
\end{lemma}
\begin{proof}
  Conditioning on the observations $s^T, a^T$ via Bayes' theorem, we
  obtain the conditional measure:
  \begin{align}
    \rbel(B \mid s^T, a^T, \cmp) &= \frac{\int_B \pbel(s^T, a^T \mid
      \rew, \cmp) \dd{\rbel}(\rew \mid \cmp)}{\int_\Rews \pbel(s^T, a^T
      \mid \rew, \cmp) \dd{\rbel}(\rew \mid \cmp)},
  \end{align}
  where $\pbel(s^T, a^T \mid \rew, \cmp) \defn \int_\Pols \Pr_{\cmp,
    \pol}(s^T, a^T) \dd{\pbel}(\pol \mid \rew, \cmp)$ is a marginal
  likelihood term.  It is easy to see via induction that:
  \begin{align}
    \Pr_{\cmp, \pol}(s^T, a^T)
    &= \prod_{t=1}^T \pol(a_t \mid s_t) \tau(s_t \mid a_{t-1},
    s_{t-1}),
    \label{eq:policy-induction}
  \end{align}
  where $\tau(s_1 \mid a_0, s_0) = \tau(s_1)$ is the initial state
  distribution.  Thus, the reward function posterior is proportional to:
  \[
    \int_B \int_\Pols \prod_{t=1}^T
    \pol(a_t | s_t) \tau(s_t | a_{t-1}, s_{t-1}) \dd{\pbel}(\pol
    | \rew, \cmp) \dd{\rbel}(\rew | \cmp).
    \]
  Note that the $\tau(s_t | a_{t-1}, s_{t-1})$ terms can be taken
  out of the integral. Since they also appear in the denominator, the
  state transition terms cancel out.
  \qed
\end{proof}

\section{Estimation}
\label{sec:model-estimation}
While it is entirely possible to assume that the agent's policy is
optimal with respect to its utility (as is done for example
in~\cite{abbeel2004apprenticeship}), our analysis can be made more
interesting by assuming otherwise. 
One simple idea is to restrict the policy space to stationary
soft-max policies:
\begin{align}
  \label{eq:softmax-prior}
  \pol_\temp(a_t \mid s_t)
  &=
  \frac{\exp(\temp \Qfs(s_t,a_t))}
  {\sum_a \exp(\temp \Qfs(s_t,a))},
\end{align}
where we assumed a finite action set for simplicity.  Then we can
define a prior on policies, given a reward function, by specifying a
prior on the inverse temperature $\eta$, such that given the reward
function and $\eta$, the policy is uniquely determined.\footnote{ Our
  framework's generality allows any functional form relating the
  agent's preferences and policies. As an example, we could define a
  prior distribution over the $\epsilon$-optimality of the chosen
  policy, without limiting ourselves to soft-max forms. This would of
  course change the details of the estimation procedure.}

For the chosen prior \eqref{eq:softmax-prior}, inference can be
performed using standard Markov chain Monte Carlo (MCMC)
methods~\cite{book:MonteCarlo}. If we can estimate the reward function
well enough, we may be able to obtain policies that surpass the
performance of the original policy $\agpol$ with respect to the
agent's reward function $\agrew$.
\begin{algorithm}[h]
  \begin{algorithmic}[1]
    \For{$k=1,\ldots$}
    \State $\tilde{\rew} \sim  \rbel(\rew \mid \cmp)$.
    \State $\tilde{\temp} \sim \GammaDist(\gammaA, \gammaB)$
    \State $\tilde{\pol} = \Softmax(\tilde{\rew}, \tilde{\temp}, \tau)$
    \State $\tilde{p} = \Pr_{\cmp, \tilde{\pol}}(s^T, a^T) / [\rbel(\rew \mid \cmp) f_{\GammaDist}(\tilde{\temp};\gammaA, \gammaB)]$.
    \WithProb{$\min\set{1, \tilde{p}/p_{(k-1)}}$}
    \State $\pol_{(k)} = \tilde{\pol}$, $\temp_{(k)} = \tilde{\temp}$, $\rew_{(k)} = \tilde{\rew}$, $p_{(k)} = \tilde{p}$.
    \Else
    \State $\pol_{(k)} = \pol_{(k-1)}$, $\temp_{(k)} = \temp_{(k-1)}$, $\rew_{(k)} = \rew_{(k-1)}$, $p_{(k)} = p_{(k-1)}$.
    \EndProb
    \EndFor
  \end{algorithmic}
  \caption{MH: Direct Metropolis-Hastings sampling from the joint
    distribution $\jbel(\pol, \rew \mid a^T, s^T)$.}
  \label{alg:mh-reward}
\end{algorithm}

\subsection{The basic model: A Metropolis-Hastings procedure}
Estimation in the basic model (Fig.~\ref{fig:basic-model}) can be
performed via a Metropolis-Hastings (MH) procedure. Recall that
performing MH to sample from some distribution with density $f(x)$
using a proposal distribution with conditional density $g(\tilde{x}
\mid x)$, has the form:
\[
x_{(k+1)} = 
\begin{cases}
  \tilde{x}, & \textrm{w.p.~} \min\set{1, \frac{f(\tilde{x})/g(\tilde{x}\mid x_{(k)})}{f(x_{(k)})/g(x_{(k)} \mid \tilde{x})}}
\\
x_{(k)}, & \textrm{otherwise}.
\end{cases}
\]

In our case, $x = (\rew, \pol)$ and $f(x) = \jbel(\rew, \pol \mid s^T,
a^T, \cmp)$.\footnote{Here we abuse notation, using $\jbel(\rew, \pol
  \mid \cdot)$ to denote the density or probability function with
  respect to a Lebesgue or counting measure associated with the
  probability measure $\jbel(B \mid \cdot)$ on subsets of $\Rews
  \times \Pols$}  We use {\em independent} proposals $g(x) =
\jbel(\rew, \pol | \cmp)$.  \iftrue As $\jbel(\rew, \pol |
s^T, a^T, \cmp) = \jbel(s^T, a^T | \rew, \pol, \cmp) \jbel(\rew,
\pol) / \jbel(s^T, a^T)$, it follows that:
\[
\frac{\jbel(\tilde{\rew}, \tilde{\pol} \mid s^T, a^T, \cmp)}
{\jbel(\rew, \pol \mid s^T, a^T, \cmp)}
 = \frac{\Pr_{\cmp,\tilde{\pol}}(s^T, a^T) \jbel(\tilde{\rew}, \tilde{\pol} \mid \cmp)}
{ \Pr_{\cmp,\pol_{(k)}}(s^T, a^T) \jbel(\rew_{(k)}, \pol_{(k)}\mid \cmp)}.
\]
This gives rise to the sampling procedure described in
Alg.~\ref{alg:mh-reward}, which uses a gamma prior for the
temperature.

\fi

\subsection{The augmented model: A hybrid Gibbs procedure}
The augmented model (Fig.~\ref{fig:augmented-model}) enables an
alternative, a two-stage hybrid Gibbs sampler, described in
Alg.~\ref{alg:gibbs-reward}.  This conditions alternatively on a
reward sequence sample $r_{(k)}^T$ and on a reward function sample
$\rew_{(k)}$ at the $k$-th iteration of the chain. Thus, we also
obtain a posterior distribution on {\em reward sequences}.

This sampler is of particular utility when the reward function prior
is conjugate to the reward distribution, in which case:
\begin{inparaenum}[(i)]
\item The reward sequence sample can be easily obtained and
\item the reward function prior can be conditioned on the
  reward sequence with a simple sufficient statistic.
\end{inparaenum}
While, sampling from the reward function posterior continues to
require MH, the resulting hybrid Gibbs sampler remains a valid
procedure~\cite{book:MonteCarlo}, which may give better results than
specifying arbitrary proposals for pure MH sampling.

As previously mentioned, the Gibbs procedure also results in a
distribution over the reward sequences observed by the agent. On the
one hand, this could be valuable in applications where the reward
sequence is the main quantity of interest. On the other hand, this has
the disadvantage of making a strong assumption about the 
distribution from which rewards are drawn.

\begin{algorithm}[h]
  \begin{algorithmic}[1]
    \For{$k=1,\ldots$}
    \State $\tilde{\rew} \sim \rbel(\rew \mid r^T_{(k-1)}, \cmp)$.
    \State $\tilde{\temp} \sim \GammaDist(\gammaA, \gammaB)$
    \State $\tilde{\pol} = \Softmax(\tilde{\rew}, \tilde{\epsilon}, \tau)$
    \State $\tilde{p} = \Pr_{\cmp, \tilde{\pol}}(s^T, a^T) / [\rbel(\rew \mid \cmp) f_{\GammaDist}(\tilde{\temp};\gammaA, \gammaB)]$.
    \WithProb{$\min\set{1, \tilde{p}/p_{(k-1)}}$}
    \State $\pol_{(k)} = \tilde{\pol}$, $\temp_{(k)} = \tilde{\temp}$, $\rew_{(k)} = \tilde{\rew}$, $p_{(k)} = \tilde{p}$.
    \Else
    \State $\pol_{(k)} = \pol_{(k-1)}$, $\temp_{(k)} = \temp_{(k-1)}$, $\rew_{(k)} = \rew_{(k-1)}$, $p_{(k)} = p_{(k-1)}$.
    \EndProb
    \State $r^{T}_{(k)} \mid s^T, a^T \sim \rew_{(k)}^T(s^T, a^T)$
    \EndFor
  \end{algorithmic}
  \caption{G-MH: Two stage Gibbs sampler with an MH step}
  \label{alg:gibbs-reward}
\end{algorithm}

\section{Related work}
\label{sec:related-work}
\subsection{Preference elicitation in user modelling}

Preference elicitation has attracted a lot of attention in the field
of user modelling and online advertising, where two main problems
exist. The first is how to {\em model} the (uncertain) preferences of
a large number of users.  The second is the problem of {\em optimal
  experiment design}
\citep[see][ch. 14]{Degroot:OptimalStatisticalDecisions} to maximise
the expected value of information through queries.  Some recent models
include: \citet{braziunas:pe-gau:aaai2006} who introduced modelling of
generalised additive utilities; \citet{chu2005preference}, who
proposed a Gaussian process prior over preferences, given a set of
instances and pairwise relations, with applications to multiclass
classification; ~\citet{Bonilla:GPPE:NIPS2010}, who generalised it to
multiple users; \citep{Guo:AISTAST:20120}, which proposed an
additively decomposable multi-attribute utility model.  Experimental
design is usually performed by approximating the intractable optimal
solution~\citep{boutilier2002pomdp,Degroot:OptimalStatisticalDecisions}.

\subsection{Inverse reinforcement learning}
As discussed in the introduction, the problems of inverse
reinforcement learning and apprenticeship learning involve an agent
acting in a {\em dynamic} environment. This makes the modelling
problem different to that of user modelling where preferences are
between static choices. Secondly, the goal is not only to determine
the preferences of the agent, but also to find a policy that would be
at least as good that of the agent with respect to the agent's own
preferences.\footnote{Interestingly, this can also be seen as the goal
  of preference elicitation when applied to multiclass
  classification~\citep[see][for example]{chu2005preference}.} Finally, the
problem of experiment design does not necessarily arise, as we do not
assume to have an influence over the agent's environment.

\subsubsection{Linear programming}
One interesting solution proposed by~\cite{Ng00algorithmsfor} is to
use a linear program in order to find a reward function that maximises
the gap between the best and second best action. Although elegant,
this approach suffers from some drawbacks.
\begin{inparaenum}[(a)]
\item A good estimate of the optimal policy must be given. This may be
  hard in cases where the demonstrating agent does not visit all of
  the states frequently.
\item In some pathological MDPs, there is no such gap. For example it
  could be that for any action $a$, there exists some other action
  $a'$ with equal value in every state. 
\end{inparaenum}

\subsubsection{Policy walk}
Our framework can be seen as a generalisation of the Bayesian approach
considered in~\cite{ramachandran51bayesian}, which does not employ a
structured prior on the rewards and policies. In fact, they implicitly
define the joint posterior over rewards and policies as:
    \begin{align*}
    \jbel(\pol, \rew \mid s^T, a^T, \cmp)
    &=
    \frac{\exp\left[\temp \sum_t \Qfs(s_t, a_t)\right] \rbel(\rew \mid \cmp)}
    {\jbel(s^T, a^T \mid \cmp)},
    \end{align*}
    which implies that the exponential term corresponds to $\rbel(s^T,
    a^T, \pol \mid \rew)$. This {\em ad hoc} choice is probably the
    weakest point in this approach.\footnote{ Although, as mentioned
      in~\cite{ramachandran51bayesian}, such a choice could be
      justifiable through a maximum entropy argument, we note that the
      maximum-entropy based approach reported
      in~\cite{ziebart:causal-entropy} does not employ the value
      function in that way.}  Rearranging, we write the denominator
    as:
    \begin{equation}
      \rbel(s^T, a^T \mid \cmp) = \int_{\mathrlap{\Rews \times \Pols}} \, \rbel(s^T, a^T \mid \pol, \rew, \cmp) \dd{\rbel}(\rew, \pol \mid \cmp),
    \end{equation}
    which is still not computable, but we can employ a
    Metropolis-Hastings step using $\rbel(\rew \mid \cmp)$ as a proposal
    distribution, and an acceptance probability of:
    \[
    \frac{\rbel(\pol, \rew \mid s^T, a^T)/\rbel(\rew)}{\rbel(\pol', \rew' \mid s^T, a^T)/\rbel(\rew')}
    =
    \frac{\exp[\temp \sum_t Q^\pol_\rew(s_t, a_t)]}
    {\exp[\temp \sum_t Q^{\pol'}_{\rew'}(s_t, a_t)]}.
    \]
    We note that in~\citep{ramachandran51bayesian}, the authors employ
    a different sampling procedure than a straightforward MH, called a
    policy grid walk.  In exploratory experiments, where we examined
    the performance of the authors' original
    method~\cite{ramachandran:pc}, we have determined that MH is
    sufficient and that the most crucial factor for this particular
    method was its initialisation: as will be also be seen in
    Sec.~\ref{sec:experiments}, we only obtained a small, but
    consistent, improvement upon the initial reward function.

\subsubsection{The maximum entropy approach.}
A maximum entropy approach is reported
in~\cite{ziebart:causal-entropy}.  Given a feature function $\Phi :
\States \times \Actions \to \Reals^n$, and a set of trajectories
$\cset{s^{T_k}_{(k)}, a^{T_k}_{(k)}}{k=1,\ldots, n}$, they obtain
features $\Phi^{T_k}_{(k)} = \left(\Phi(s_{i,(k)},
  a_{i,(k)})\right)_{i=1}^{T_k}$.  They show that given empirical
constraints $\E_{\theta, \cmp} \Phi^{T_k} = \hE \Phi^{T_k}$, where
$\hE \Phi^T = \frac{1}{n} \sum_{k=1}^n \Phi^{T_k}_{(k)}$ is the
empirical feature expectation, one can obtain a maximum entropy
distribution for actions of the form $\Pr_{\theta}(a_t \mid s_t)
\propto e^{\theta' \Phi(s_t, a_t)}$. If $\Phi$ is the identity, then
$\theta$ can be seen as a scaled state-action value function.

In general, maximum entropy approaches have good minimax
guarantees~\citep{grunwald:max-ent:aos:2004}. Consequently, the
estimated policy is guaranteed to be close to the agent's.  However,
at best, by bounding the error in the policy, one obtains a two-sided
high probability bound on the relative loss. Thus, one is almost
certain to perform neither much better, nor much worse that the
demonstrator.

\subsubsection{Game theoretic approach}
An interesting game theoretic approach was suggested
by~\cite{syed-schapire:game-theory:nips} for apprenticeship
learning. This also only requires statistics of observed features,
similarly to the maximum entropy approach. The main idea is to find
the solution to a game matrix with a number of rows equal to the
number of possible policies, which, although large, can be solved
efficiently by an exponential weighting algorithm. The method is
particularly notable for being (as far as we are aware of) the only
one with a high-probability upper bound on the loss relative to the
demonstrating agent and no corresponding lower bound. Thus, this
method may in principle lead to a significant improvement over the
demonstrator. Unfortunately, as far as we are aware of, sufficient
conditions for this to occur are not known at the moment.  In more
recent work~\citep{apprenticeship:classification}, the authors have
also made an interesting link between the error of a classifier trying
to imitate the expert's behaviour and the performance of the imitating
policy, when the demonstrator is nearly optimal.

\begin{figure}[!h]
\begin{center}
  \subfigure[Random MDP]{
      \psfrag{Softmax temperature of demonstrator}[B][B][1.0][0]{$\eta$}
      \psfrag{Loss with greedy policy}[B][B][1.0][0]{$\loss$}
      \includegraphics[width=0.475\columnwidth]{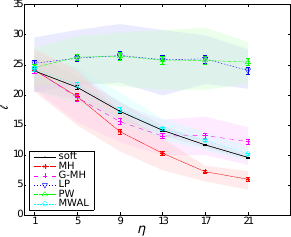}
    }
    \subfigure[Random Maze]{
      \psfrag{Softmax temperature of demonstrator}[B][B][1.0][0]{$\eta$}
      \psfrag{Loss with greedy policy}[B][B][1.0][0]{$\loss$}
      \includegraphics[width=0.475\columnwidth]{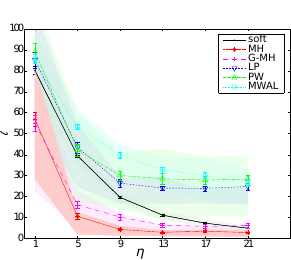}
    }
    \caption{Total loss $\loss$ with respect to the optimal policy, as
      a function of the inverse temperature $\eta$ of the softmax
      policy of the demonstrator for (a) the Random MDP and (b) the
      Random Maze tasks, averaged over 100 runs.  The shaded areas
      indicate the 80\% percentile region, while the error bars the
      standard error.}
\label{fig:loss-temperature-greedy}
\end{center}
\end{figure}

\section{Experiments}
\label{sec:experiments}
\subsection{Domains}
We compare the proposed algorithms on two different domains, namely on
random MDPs and random maze tasks.  The {\em Random MDP} task is a
discrete-state MDP, with four actions, such that each leads to a
different, but possibly overlapping, quarter of the state
set.\footnote{ The transition matrix of the MDPs was chosen so that
  the MDP was communicating (c.f.~\citep{Puterman:MDP:1994}) and so
  that each individual action from any state results in a transition
  to approximately a quarter of all available states (with the
  destination states arrival probabilities being uniformly selected
  and the non-destination states arrival probabilities being set to
  zero).}  The reward function is drawn from a Beta-product hyperprior
with parameters $\alpha_{i}$ and $\beta_{i}$, where the index $i$ is
over all state-action pairs.  This defines a distribution over the
parameters $p_{i}$ of the Bernoulli distribution determining the
probability of the agent of obtaining a reward when carrying out an
action $a$ in a particular state $s$.

For the {\em Random Maze} tasks we constructed {\em
  planar grid} mazes of different sizes, with four actions at each
state, in which the agent has a probability of $0.7$ to succeed with
the current action and is otherwise moved to one of the adjacent
states randomly. These mazes are also randomly generated, with the
rewards function being drawn from the same prior. The maze structure
is sampled by randomly filling a grid with walls through a
product-Bernoulli distribution with parameter $1/4$, and then rejecting any 
mazes with a number of obstacles higher than $|\States|/4$.

\subsection{Algorithms, priors and parameters}

We compared our methodology, using the basic ({\bf MH}) and the
augmented ({\bf G-MH}) model, to three previous approaches. The linear
programming ({\bf LP}) based approach~\citep{Ng00algorithmsfor}, the
game-theoretic approach ({\bf
  MWAL})~\citep{syed-schapire:game-theory:nips} and finally, the
Bayesian inverse reinforcement learning method ({\bf PW}) suggested
in~\citep{ramachandran51bayesian}.  In all cases, each demonstration
was a $T$-long trajectory $s^T, a^T$, provided by a demonstrator
employing a softmax policy with respect to the optimal value function.

All algorithms have some parameters that must be selected.  Since our
methodology employs MCMC the sampling parameters must be chosen so
that convergence is ensured. We found that $10^4$ samples from the
chain were sufficient, for both the MH and hybrid Gibbs (G-MH)
sampler, with $2000$ steps used as burn-in, for both tasks. In both
cases, we used a gamma prior $\GammaDist(1,1)$ for the inverse
temperature parameter $\eta$ and a product-beta prior
$\Beta^{|S|}(1,1)$ for the reward function. Since the beta is
conjugate to the Bernoulli, this is what we used for the reward
sequence sampling in the G-MH sampler. Accordingly, the conditioning
performed in step $11$ of G-MH is closed-form.

For {\bf PW}, we used a MH sampler seeded with the solution found
by~\cite{Ng00algorithmsfor}, as suggested by~\cite{ramachandran:pc}
and by our own preliminary experiments. Other initialisations, such as
sampling from the prior, generally produced worse results. In
addition, we did not find any improvement by discretising the sampling
space. We also verified that the same number of samples used in our
case was also sufficient for this method to converge.

The linear-programming ({\bf LP}) based inverse reinforcement learning
algorithm by~\citet{Ng00algorithmsfor} requires the actual agent
policy as input. For the {\em random MDP} domain, we used the maximum
likelihood estimate.  For the maze domain, we used a Laplace-smoothed
estimate (a product-Dirichlet prior with parameters equal to 1)
instead, as this was more stable.

Finally, we examined the {\bf MWAL} algorithm
of~\citet{syed-schapire:game-theory:nips}. This requires the
cumulative discounted feature expectation as input, for appropriately
defined features. Since we had discrete environments, we used the
state occupancy as a feature. The feature expectations can be
calculated empirically, but we obtained better performance by first
computeing the transition probabilities of the Markov chain induced by
the maximum likelihood (or Laplace-smoothed) policy and then
calculating the expectation of these features given this chain. We set
all accuracy parameters of this algorithm to $10^{-3}$, which was
sufficient for a robust behaviour.

\subsection{Performance measure}
In order to measure performance, we plot the $L_1$ loss\footnote{This
  loss can be seen as a scaled version of the expected loss under a
  uniform state distribution and is a bound on the $L_\infty$ loss.
  The other natural choice of the optimal policy stationary state
  distribution is problematic for non-ergodic MDPs.}  of the value
function of each policy relative to the optimal policy with respect to
the agent's utility:
\begin{equation}
\loss(\pi) \defn \sum_{s \in \States} \Vfs(s) - \Vfp(s),
\label{eq:loss}
\end{equation}
where $\Vfs(s) \defn \max_a \Qfs(s, a)$ and $\Vfp(s) \defn \E_\pi
\Qfp(s, a)$.  

In all cases, we average over $100$ experiments on an equal number of
randomly generated environments $\mdp_1, \mdp_2, \ldots$. For the
$i$-th experiment, we generate a $T$-step-long demonstration $D_i =
(s^T, a^T)$ via an agent employing a softmax policy. The same
demonstration is used across all methods to reduce variance.  In
addition to the empirical mean of the loss, we use shaded regions to
show $80\%$ percentile across trials and error bars to display the
standard error.

\begin{figure}[h!]
  \begin{center}
    \subfigure[Effect of amount of data]{
      \psfrag{Loss with greedy policy}[B][B][1.0][0]{$\loss$}
      \psfrag{Number of demonstrations}[B][B][1.0][0]{$T$}
      \includegraphics[width=0.475\columnwidth]{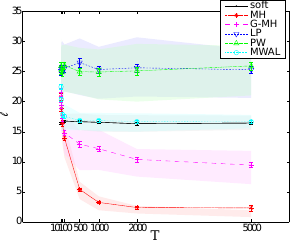}
      \label{fig:loss-rmdp-data-greedy}
    }
    \subfigure[Effect of environment size]{
      \psfrag{Number of states}[B][B][1.0][0]{$|\CS|$}
      \psfrag{Loss with greedy policy}[B][B][1.0][0]{$\loss$}
      \includegraphics[width=0.45\columnwidth]{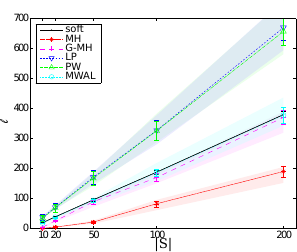}
      \label{fig:loss-rmdp-states-greedy}
    }
    \caption{Total loss $\loss$ with respect to the optimal policy, in
      the Random MDP task.  Figure~\ref{fig:loss-rmdp-data-greedy}
      shows how performance improves as a function of the length
      $T$. of the demonstrated sequence.
      Figure~\ref{fig:loss-rmdp-states-greedy} shows the effect of the
      number of states $|\CS|$ of the underlying MDP.  All quantities
      are averaged over 100 runs. The shaded areas indicate the 80\%
      percentile region, while the error bars the standard error.}
  \end{center}
\end{figure}

\subsection{Results}
\label{sec:results}
We consider the loss of five different policies. The first, {\bf
  soft}, is the policy of the demonstrating agent itself. The second,
{\bf MH}, is the Metropolis-Hastings procedure defined in
Alg.~\ref{alg:mh-reward}, while {\bf G-MH} is the hybrid Gibbs
procedure from Alg.~\ref{alg:gibbs-reward}. We also consider the loss
of our implementations of Linear Programming ({\bf LP}), Policy Walk
({\bf PW}), and {\bf MWAL}, as summarised in
Sec.~\ref{sec:related-work}.

We first examined the loss of greedy policies,\footnote{Experiments
  with non-greedy policies (not shown) produced generally worse
  results.} derived from the estimated reward function, as the
demonstrating agent becomes
greedier. Figure~\ref{fig:loss-temperature-greedy} shows results for
the two different domains. It is easy to see that the {\bf MH} sampler
significantly outperforms the demonstrator, even when the latter is
near-optimal. While the hybrid {\bf Gibbs} sampler's performance lies
between that of the demonstrator and the {\bf MH} sampler, it also
estimates a distribution over reward sequences as a side-effect. Thus,
it could be of further value where estimation of reward sequences is
important.  We observed that the performance of the baseline methods
is generally inferior, though nevertheless the {\bf MWAL} algorithm
tracks the demonstrator's performance closely.

This suboptimal performance of the baseline methods in the {\em Random
  MDP} setting cannot be attributed to poor estimation of the
demonstrated policy, as can clearly be seen in
Figure~\ref{fig:loss-rmdp-data-greedy}, which shows the loss of the greedy
policy derived from each method as the amount of data increases. While
the proposed samplers improve significantly as observations
accumulate, this effect is smaller in the baseline methods we compared
against.  As a final test, we plot the relative loss in the {\em
  Random MDP} as the number of states increases in
Figure~\ref{fig:loss-rmdp-states-greedy}. We can see that the relative
performance of methods is invariant to the size of the state space for
this problem.

Overall, we observed the basic model ({\bf MH}) consistently
outperforms\footnote{It was pointed out by the anonymous reviewers,
  that the loss we used may be biased. Indeed, a metric defined over
  some other state distribution, could give different
  rankings. However, after looking at the results carefully we
  determined that the policies obtained via the MH sampler were
  strictly dominating.}
the agent in all settings. The augmented model ({\bf G-MH}), while
sometimes outperforming the demonstrator, is not as consistent.
Presumably, this is due to the joint estimation of the reward
sequence.  Finally, the other methods under consideration on average
do not improve upon the initial policy and can be, in a large number
of cases, significantly worse. For the linear programming inverse RL
method, perhaps this can be attributed to implicit assumptions about
the MDP and the optimality of the given policy.  For the policy walk
inverse RL method, our belief is that its suboptimal performance is
due to the very restrictive prior it uses. Finally, the performance of
the game theoretic approach is slightly disappointing. Although it is
much more robust than the other two baseline approaches, it never
outperforms the demonstrator, even thought technically this is
possible.  One possible explanation is that since this approach is
worst-case by construction, it results in overly conservative
policies.

\section{Discussion}
\label{sec:discussion}

We introduced a unified framework of preference elicitation and
inverse reinforcement learning, presented two statistical inference
models, with two corresponding sampling procedures for estimation. Our
framework is flexible enough to allow using alternative priors on the
form of the policy and of the agent's preferences, although that would
require adjusting the sampling procedures. In experiments, we showed
that for a particular choice of policy prior, closely corresponding to
previous approaches, our samplers can outperform not only other
well-known inverse reinforcement learning algorithms, but the
demonstrating agent as well.

The simplest extension, which we have already alluded to, is the
estimation of the discount factor, for which we have obtained
promising results in preliminary experiments.  A slightly harder
generalisation occurs when the environment is not known to us. This is
not due to difficulties in inference, since in many cases a posterior
distribution over $\MDPs$ is not hard to maintain (see for
example~\citep{duff2002olc,poupart2006asd}). However, computing the
optimal policy given a belief over MDPs is harder~\citep{duff2002olc},
even if we limit ourselves to stationary
policies~\citep{DBLP:journals/jmlr/FurmstonB10}.  We would also like
to consider more types of preference and policy priors. Firstly, the
use of spatial priors for the reward function, which would be
necessary for large or continuous environments.  Secondly, the use of
alternative priors on the demonstrator's policy. 

The generality of the framework allows us to formulate different
preference elicitation problems than those directly tied to
reinforcement learning. For example, it is possible to estimate
utilities that are not additive functions of some latent rewards.
This does not appear to be easily achievable through the extension of
other inverse reinforcement learning algorithms. It would be
interesting to examine this in future work.

Another promising direction, which we have already investigated to
some degree~\cite{MTIRL}, is to extend the framework to a fully
hierarchical model, with a hyperprior on reward functions. This would
be particularly useful for modelling a {\em population} of
agents. Consequently, it would have direct applications on the
statistical analysis of behavioural experiments.

Finally, although in this paper we have not considered the problem of
{\em experimental design} for preference elicitation (i.e. {\em
  active} preference elicitation), we believe is a very interesting
direction.  In addition, it has many applications, such as online
advertising and the automated optimal design of behavioural
experiments. It is our opinion that a more effective preference
elicitation procedure such as the one presented in this paper is
essential for the complex planning task that experimental design
is. Consequently, we hope that researchers in that area will find our
methods useful.

\subsubsection*{Acknowledgements}
Many thanks to the anonymous reviewers for their comments and
suggestions. This work was partially supported by the BMBF Project
"Bernstein Fokus: Neurotechnologie Frankfurt, FKZ 01GQ0840'', the
EU-Project IM-CLeVeR, FP7-ICT-IP-231722, and the Marie Curie Project
ESDEMUU, Grant Number 237816.

\bibliographystyle{plainnat}
\bibliography{inverse_rl}

\end{document}

%% file: preamble.tex
\usepackage{enumerate}
\usepackage{amsmath}
\usepackage{amsfonts}
\usepackage{amssymb}
\usepackage{amsbsy}
\usepackage{dsfont}
\usepackage{theorem}
\usepackage{algorithm}
\usepackage{algorithmicx}
\usepackage{algpseudocode}
\usepackage{mathrsfs}
\usepackage{paralist}
\usepackage{epsfig}
\usepackage{subfigure}
\usepackage{makeidx} 
\usepackage{color}
\usepackage{pstricks}
\usepackage{pst-node}
\usepackage{multido} 
\usepackage{wrapfig} 
\usepackage{tikz}
\usepackage[numbers]{natbib}

\numberwithin{equation}{section} 

\theoremstyle{plain} \newtheorem{assumption}{Assumption}

\newcommand \E {\mathop{\mbox{\ensuremath{\mathbb{E}}}}\nolimits}
\newcommand \hE {\hat{\mathop{\mbox{\ensuremath{\mathbb{E}}}}\nolimits}}
\renewcommand \Pr {\mathop{\mbox{\ensuremath{\mathbb{P}}}}\nolimits}

\newcommand{\set}[1]{\left\{\, #1 \,\right\} }
\newcommand{\cset}[2]{\left\{\, #1 \mathrel{:} #2 \,\right\} }

\newcommand\Reals {{\mathds{R}}}

\newcommand \CA {{\mathcal{A}}}

\newcommand \CM {{\mathcal{M}}}

\newcommand \CP {{\mathcal{P}}}

\newcommand \CR {{\mathcal{R}}}
\newcommand \CS {{\mathcal{S}}}
\newcommand \CT {{\mathcal{T}}}

\newcommand \defn {\mathrel{\triangleq}}

\newcommand \Qf {Q_\mdp}
\newcommand \Qfp {Q_\mdp^\pol}
\newcommand \Qfs {Q_\mdp^*}

\newcommand \Vfp {V_\mdp^\pol}
\newcommand \Vfs {V_\mdp^*}

\DeclareMathAlphabet{\mathpzc}{OT1}{pzc}{m}{it}

\newcommand \Beta {\mathop{\mathpzc{Beta}}\nolimits}
\newcommand \GammaDist {\mathop{\mathpzc{Gamma}}\nolimits}
\newcommand \Softmax{\mathop{\mathpzc{Softmax}}\nolimits}

\newcommand \Actions {\CA}
\newcommand \States {\CS}
\newcommand \Trans {\CT}
\newcommand \tran[2] {\tau(#1 \mid #2)}
\newcommand \rew {\rho}
\newcommand \Rews {\CR}
\newcommand \disc {\gamma}
\newcommand \mdp {\mu}
\newcommand \cmp {\nu}
\newcommand \MDPs {\CM}
\newcommand \pol {\pi}
\newcommand \Pols {\CP}

\newcommand \agrew {\rew_{\mathrm{a}}}
\newcommand \agpol {\pol_{\mathrm{a}}}

\newcommand \temp {\eta}
\newcommand \gammaA {\zeta}
\newcommand \gammaB {\theta}

\newcommand \jbel {\phi}

\newcommand \rbel {\xi}
\newcommand \pbel {\psi}

\newcommand \loss {\ell}

\newcommand \Ase {A^*_\epsilon}

\newcommand\dd{\,\mathrm{d}}

\algblockdefx[WithProb]{WithProb}{EndProb}
[1]{{\bf w.p.} #1 {\bf do}}{\bf done}
\algcblock{WithProb}{Else}{EndProb}

\def\clap#1{\hbox to 0pt{\hss#1\hss}}

\def\mathrlap{\mathpalette\mathrlapinternal}

\def\mathrlapinternal#1#2{%
           \rlap{$\mathsurround=0pt#1{#2}$}}

\tikzstyle{place}=[circle,draw=black,draw=blue!50,fill=blue!20,inner sep=0mm, minimum size=6mm]
\tikzstyle{hidden}=[circle,draw=black,draw=blue!50,fill=blue!01,inner sep=0mm, minimum size=6mm]
\tikzstyle{observed}=[circle,draw=black,draw=blue!50,fill=blue!10,inner sep=0mm, minimum size=6mm]
\tikzstyle{transition}=[rectangle,draw=black!50,fill=black!20,thick]